# Convex Coding

**David M. Bradley**
Robotics Institute
Carnegie Mellon University
Pittsburgh, PA 15213

**J. Andrew Bagnell**
Robotics Institute
Carnegie Mellon University
Pittsburgh, PA 15213

## Abstract

Inspired by recent work on convex formulations of clustering (Lashkari & Golland, 2008; Nowozin & Bakir, 2008) we investigate a new formulation of the *Sparse Coding Problem* (Olshausen & Field, 1997). In sparse coding we attempt to simultaneously represent a sequence of data-vectors sparsely (i.e. *sparse approximation* (Tropp et al., 2006)) in terms of a "code" defined by a set of basis elements, while also finding a code that enables such an approximation. As existing alternating optimization procedures for sparse coding are theoretically prone to severe local minima problems, we propose a convex relaxation of the sparse coding problem and derive a boosting-style algorithm, that (Nowozin & Bakir, 2008) serves as a convex "master problem" which calls a (potentially non-convex) sub-problem to identify the next code element to add. Finally, we demonstrate the properties of our boosted coding algorithm on an image denoising task.

## 1 Introduction

A crucial part of many machine learning applications is representing the raw input in terms of a "code", i.e. a set of features which captures the aspects of the input examples that are relevant to prediction. Unsupervised techniques such as clustering and sparse coding (Olshausen & Field, 1997) learn codes which capture the structure of unlabeled data, and have shown to be useful for a variety of machine learning problems (Raina et al., 2007; Bradley & Bagnell, 2009b; Mairal et al., 2009). In these techniques the input is represented as a combination of the features (also known as basis vectors or dictionary elements) that make up the code. The traditional approach to clustering and coding problems is to alternate between optimizing over the elements of the code and the combinations of elements used to represent the raw input. However, this "alternating optimization" approach is non-convex with many local minima, leading to recent work on alternative, convex versions of clustering and coding (Lashkari & Golland, 2008; Nowozin & Bakir, 2008; Bach et al., 2008).

This work adds several main contributions: we present a regularization function based on *compositional norms* that implements a convex version of sparse coding, we derive the *Fenchel conjugate* of these compositional norms, and we show how Fenchel conjugates can be used to construct an efficient boosting algorithms for convex (including non-differentiable) regularization functions.

Clustering and coding can be viewed as matrix factorization problems (Ding et al., 2005; Singh & Gordon, 2008), which seek to approximate a set of input signals $X \approx f(BW)$ with the product of a dictionary matrix $B$ a coefficient (or weight) matrix $W$ and an elementwise transfer function $f$. When $B$ is known and fixed and a regularization function is used to encourage $W$ to be "sparse", this is the sparse approximation technique developed in engineering and the sciences. Sparse approximation relies on an *optimization* algorithm to infer the Maximum A-Posteriori (MAP) weights $\hat{W}$ that best reconstruct the signal. In this notation, each input signal forms a column of an input matrix $X$, and is generated by multiplying the dictionary (or basis matrix) $B$ by a column from $W$, and (optionally) applying a transfer function $f$. This relationship is only approximate, as the input data is assumed to be corrupted by random noise. Priors which produce sparse solutions for $W$, especially $L_1$ regularization, have gained attention because of their usefulness in ill-posed engineering problems (Tropp, 2006), elucidating neuro-biological phenomena, (Olshausen & Field, 1997; Karklin & Lewicki, 2005),face recognition (Wright et al., 2009), and semi-supervised and transfer learning (Raina et al., 2007; Bradley & Bagnell, 2009b; Mairal et al., 2009).

Sparse coding (Olshausen & Field, 1997) extends

sparse approximation by also learning the basis matrix $B$. Clustering can also be viewed as a restricted form of the coding matrix factorization problem (Ding et al., 2005); a special case of coding where the basis vectors are the cluster centroids and the $W$ matrix is the cluster membership of each example. Recently (Lashkari & Golland, 2008) showed that the clustering problem can be made convex by considering a fixed set of possible cluster centroids. Exemplars are a natural choice for these candidate cluster centroids, but (Nowozin & Bakir, 2008) show that for some problems this can be overly restrictive, and better results can be achieved by defining the problem in terms of a convex "master" problem, and a subproblem where new centroid candidates are generated.

We extend this "convex clustering" approach to the coding setting. Starting from a convex but intractable version of sparse coding in Section 2, we derive in Section 3 a boosting-style approach that is convex except for a subproblem. In Section 4 we give an efficient algorithm for solving the subproblem that will only improve on a fully convex exemplar-based approach, which is demonstrated on an image denoising task in Section 5. A similar convex formulation of sparse coding was independently developed by (Bach et al., 2008), who present an interesting optimization approach based on convex relaxations. Our work complements theirs by providing a novel optimization strategy applicable to all convex regularization functions.

## 2 Sparse Coding

### 2.1 Notation

Uppercase letters, $X$, denote matrices and lowercase letters, $x$, denote vectors. For matrices, superscripts and subscripts denote rows and columns respectively. $X_j$ is the jth column of $X$, $X^i$ is the ith row of $X$, and $X^i_j$ is the element in the ith row and jth column. Elements of vectors are indicated by subscripts, $x_j$. $X^T$ is the transpose of matrix $X$. The Fenchel conjugate of a function $f$ is denoted by $f^*$. The notation $(x)_+$ means the larger of 0 or $x$.

### 2.2 Generative Model

From a probabilistic viewpoint, sparse coding fits a generative model (1) to unlabeled data, which factorizes a matrix of input examples, $X \in \Re^{m \times n}$, in terms of latent variable matrices $B \in \Re^{m \times d}$ and $W \in \Re^{d \times n}$. The matrix $B$ is referred to as the basis, code, or dictionary, and is shared across all $n$ examples (columns of $X$). Given $B$ the examples are assumed to be independent of each other. The matrix $W$ is commonly referred to as the coefficients or the activations of the basis vectors.

$$P(X) = \int_B \int_W P(X|BW)P(W)P(B) dW dB \quad (1)$$

Applying the Maximum A Posteriori (MAP) approximation replaces the integration over $W$ and $B$ in (1) with its maximum value $P(X|\hat{B}\hat{W})P(\hat{W})P(\hat{B})$, where the values of the latent variables at the maximum, $\hat{W}$ and $\hat{B}$, are referred to as the MAP estimates. Finding $\hat{W}$ given $B$ is an approximation problem; solving for $\hat{W}$ and $\hat{B}$ simultaneously over a set of examples is a coding problem.

We will focus on the case examined in (Lee et al., 2007) where the input $X$ is assumed to be the matrix product $BW$ corrupted by additive i.i.d Gaussian noise on each element, the prior $P(W)$ is assumed to be Laplacian with mean zero, and the the columns of the basis matrix are constrained to unit length. For numerical stability we minimize the negative log probability instead of maximizing (1):

$$\hat{W}, \hat{B} = \arg\min_{W,B} \|BW - X\|_F^2 + \lambda \|W\|_1$$
$$s.t. \quad \|B_i\|_2 = 1, \ \forall i. \quad (2)$$

The $L_1$-norm[1], is a common choice for the regularization function $\Phi(W)$ because it tends to produce $\hat{W}$ which are "sparse"–contain a small number of non-zero elements–even when the basis matrix has infinitely many columns (Bengio et al., 2006). This preference has been shown to be useful for applications such as prediction (Raina et al., 2007; Bradley & Bagnell, 2009b; Mairal et al., 2009) and denoising of images and video (Mairal et al., 2008).

### 2.3 Convex Relaxation

This formulation, and similar variants, is commonly solved by alternating between optimization over $W$ and optimization over $B$, as both problems are convex when the other matrix is constant. However, a common objection to this approach is that the joint optimization problem is non-convex.

As noted by (Bengio et al., 2006) for the related problem of learning neural networks, the non-convexity can be removed if $B$ is a fixed, infinite basis matrix containing all unit-length vectors as columns, and the optimization is only with respect to $W$. They go on to show that if $L_1$ regularization is placed on $W$, it will have optimal solutions with only a finite number of non-zero weights, even if the number of basis vectors is infinite. Hence, the matrix $B$ in (2) can be interpreted as the small set of basis vectors that have non-zero weight in $W$, and the fixed number of columns of $B$ can be written as a *compositional norm* constraint on W.

---
[1] The $L_p$ norm of a vector $x$ is: $\|x\|_p = \left(\sum_i |x|_i^p\right)^{1/p}$, for $p \geq 1$

### 2.3.1 Compositional Norms

A compositional norm is a norm composed of norms[2] over disjoint sets of variables (Bradley & Bagnell, 2009a). A useful and notationally convenient example of a compositional norm is a block norm[3]. Define a block norm of the matrix $W$, $\|W\|_{p,q}$ to be the $L_q$ norm of the $L_p$ norms of every row $W^i$:

$$L_{p,q}(W) = \|W\|_{p,q} = \left( \sum_i \left( \sum_j |W_j^i|^p \right)^{\frac{q}{p}} \right)^{\frac{1}{q}} \quad (3)$$

Since in our setting $W$ is defined so that each row corresponds to a basis vector and each column corresponds to an example, a block norm can encourage all examples to use a subset of the basis vectors. For instance, the fixed size of $B$ in (2) is equivalent to a hard constraint on $W$ in terms of the non-convex $L_{2,0}$ block semi-norm. $L_{2,0}$ is the $L_0$ semi-norm[4] of the $L_2$ norm of each row of $W$, which counts how many basis vectors have non-zero entries in $W$.

Our convex coding formulation relaxes this non-convex $L_{2,0}$ constraint by substituting regularization with the convex (but still sparse) $L_{2,1}$ constraint:

$$\arg\min_W \frac{1}{2}\|BW - X\|_F^2 + \lambda \left( \frac{1}{2}\|W\|_{2,1}^2 + \frac{\gamma}{2}\|W\|_1^2 \right) \quad (4)$$

The fact that the norms are squared in (4) will be mathematically convenient later, and is equivalent to scaling the regularization constant[5].

The $L_{2,1}$ block norm has been advocated recently as a regularization function for *multi-task learning* (Obozinski et al., 2006; Tropp et al., 2006), and the combination of the $L_{2,1}$ block norm with the $L_1$ norm was independently used for sparse coding by (Bach et al., 2008), although presented quite differently in terms of decomposition norms. Their work provides an interesting alternative framework and optimization strategy to the boosting approach presented here.

## 3 A Boosting Approach to Coding

With a finite basis matrix $B$, (4) can be solved directly with various convex optimization techniques, including, e.g., subgradient descent. (Zinkevich, 2003). The regularization term effectively "selects" a small active set from the full basis (by encouraging multiple examples to share the same bases) to form the coded representation of the input. However, by exploiting properties of the $L_{2,1}^2 + \gamma L_1^2$ regularization function it is also possible to handle an infinitely large $B$. We show how to solve (4) with an efficient boosting algorithm (Algorithm 1) that adds one new basis vector to the active basis matrix in each step.

This approach is motivated by the view of boosting as functional gradient descent in the space of weak learners (Mason et al., 2000; Friedman, 2001). In our case, each weak learner is a vector with unit length. Each boosting step, attempts to maximize the correlation between the negative loss gradient and a "small" change in $W$, as measured by a regularization function $\Phi(W)$. For infinite sets of potential basis vectors, the maximization at each step of this boosting approach is a non-convex sub-problem which must be solved (or approximated) by an oracle (Algorithm 2).

### 3.1 Fenchel Conjugate

A useful tool in our analysis will be the the Fenchel-Legendre conjugate (5), also known as the conjugate function (Boyd & Vandenberghe, 2004), which generalizes Legendre duality to include non-differentiable functions:

$$f^*(z) = \sup_x \left\{ x^T z - f(x) \right\}. \quad (5)$$

$f^*(z)$ is the conjugate of the function $f(x)$, and the variable $z$ is the dual variable of $x$. When the supremum in (5) is achieved, every maximal value $\hat{x}$ of (5) is a subgradient[6] with respect to $z$ of the conjugate function $f^*(z)$:

$$\frac{\partial f^*(z)}{\partial z} = \hat{x} = \arg\max_x \left\{ x^T z - f(x) \right\}. \quad (6)$$

### 3.2 Boosting With Fenchel Conjugates

Each step of a gradient boosting style algorithm (Mason et al., 2000; Friedman, 2001) seeks a descent direction which provides the greatest reduction of loss for a small increase in the regularization function. If $w$ is a vector of weights over all of the possible weak learners, we wish to find the step $\Delta \hat{w}$ that is both maximally correlated with the negative loss gradient $\nabla L(w)$, and smaller than $\epsilon$ as measured by the regularization function $\Phi(w)$:

$$\Delta \hat{w} = \arg\max_{\Phi(\Delta w) \leq \epsilon} -\nabla L(w)^T \Delta w. \quad (7)$$

---

[2]The component norms of a compositional norm can also be compositional norms, allowing heirarchical arrangements of three or more norms.

[3]It can be easily verified that (3) satisfies the definition of a norm.

[4]For $0 \leq p < 1$, the $L_p$ norm of a vector $x$ is redefined as: $\|x\|_p = \sum_i |x_i|^p$

[5]At the minimum $\hat{W}$ of (4), $\lambda \|\hat{W}\|_{2,1}^2 = \lambda' \|W\|_{2,1}$, where $\lambda' = \lambda \|\hat{W}\|_{2,1}$ is the original regularization constant scaled by the $L_{2,1}$-norm of $\hat{W}$.

[6]A vector $\phi \in R^n$ is a subgradient, $\phi \in \partial_x f(x)$, of a function $f : R^2 \to (-\infty, \infty]$ at $x \in R^n$ if $\phi^t y \leq f(x + y) - f(x)$, $\forall y \in R^n$. Consider the function $f = \max(x_1, x_2)$. If $x_1 > x_2$, then there is a unique gradient $\frac{\partial f}{\partial x} = [1\ 0]$. However, if $x_1 = x_2$, then $[1\ 0]$ and $[0\ 1]$ (and any convex combination of the two) are subgradients. $\partial_x f(x)$ is the set of all subgradients.

This section will show that if the regularization function $\Phi$ is convex, defined on $\Re^d$, and the constraint is strictly feasible[7], every subgradient of its Fenchel conjugate evaluated on the loss gradient, $\Delta \hat{w} \in \partial \Phi^*(-\frac{1}{\lambda}\nabla L(w))$, is an optimal boosting step according to (7). This provides a useful method for constructing boosting algorithms for a wide class of regularization functions, and we will apply it to (4).

First we upper bound (7) by the minimum of the unconstrained Lagrange dual function, assuming the step size constraint is feasible (i.e. there exists a $\Delta w$ such that $\Phi(\Delta w) \leq \epsilon$):

$$\min_{\lambda \geq 0} \max_{\Delta w} \; -\nabla L(w)^T \Delta w - \lambda \left( \Phi(\Delta w) - \epsilon \right). \quad (8)$$

If $\Phi$ meets the conditions above, then the pair $(\Delta \hat{w}, \hat{\lambda})$ that optimizes the upper bound (8), will also be optimal for the primal (7), iff the KKT conditions (generalized to subdifferential functions) are satisfied (Borwein & Lewis, 2006). In this case the conditions state that the negative loss gradient must be parallel to a subgradient of the regularization function (9), and all of the active constraints on $\Delta \hat{w}$ must be tight (10):

$$-\nabla L(w) \in \hat{\lambda} \partial \Phi(\Delta \hat{w}) \quad (9)$$
$$\hat{\lambda} \left( \Phi(\Delta \hat{w}) - \epsilon \right) = 0. \quad (10)$$

Since $\nabla L(w)^T \Delta w$ is a linear function of $\Delta w$ and there is only one constraint, it must be active (i.e. $\lambda > 0$) whenever $\nabla L(w) \neq 0$. Since boosting would stop if $\nabla L(w) = 0$, dividing (8) by $\lambda$ and adding $\epsilon$ does not change the optimal values of $(\Delta \hat{w}, \hat{\lambda})$, and produces the definition of the Fenchel conjugate of the regularization function, with the dual variable $z = -\frac{1}{\lambda}\nabla L(w)$:

$$\max_{\Delta w} \left\{ -\frac{1}{\lambda}\nabla L(w)^T \Delta w - \Phi(\Delta w) \right\} = \Phi^*(z). \quad (11)$$

Further, (6) means that all subgradients $\Delta \hat{w} \in \partial_z \Phi^*(z)$ are boosting steps which will optimize (7). Hence for convex functions we find a boosting update rule:

$$\Delta \hat{w} = \partial \Phi^* (-\frac{1}{\lambda}\nabla L(w)) \quad (12)$$

that is a natural subgradient generalization of the mirror-descent rule for Legendre regularization functions (Cesa-Bianchi & Lugosi, 2006):

$$\Delta \hat{w} = \nabla \Phi^* (-\frac{1}{\lambda}\nabla L(w)). \quad (13)$$

By extending the mirror-descent rule to convex but non-differentiable reglarization functions, we gain the ability to use mirror descent to optimize over infinite-dimensional spaces, while only ever storing a finite

---

[7]Slater's Condition, which in this case means $\Phi(\Delta w) < \epsilon$ for some vector $\Delta w \in \Re^d$

---

**Algorithm 1** Boosted Coding

**Input:** Data matrix $X \in \Re^{m \times n}$, scalars $d, \lambda \in \Re^+$, convex functions $L(BW, X)$, $\Phi(W)$, and a function $b = \text{oracle}(-1/\lambda \nabla L(BW, X))$ which returns a new basis vector $b$ corresponding to a non-zero row of $\Delta w$. (12).
**Output:** Active basis matrix $B \in \Re^{m \times d}$ and coefficients $W \in \Re^{d \times n}$.
**Initialize:** $W = 0^{d \times n}$, $B = 0^{m \times d}$.
**for** $t = 1$ to $d$ **do**
  Add a new basis vector with zero weight:
  **1:** $B_t = \text{oracle}(-1/\lambda \nabla L(BW, X))$, $W^t = \vec{0}^T$
  Optimize W:
  **2:** $W = \arg\min_W L(BW, X) + \lambda \Phi(W)$
  **if** $\|W^t\|_2 = 0$ **then**
    return
  **end if**
**end for**

---

number of non-zero entries in $w$. The key is to employ regularization functions like $L_1$ that can have extremely sparse conjugate subgradients, and to find a computational trick (oracle) that can compute a subgradient (12) without ever explicitly computing the full gradient of the loss (which will be infinitely large for an infinite set of possible basis vectors).

In the following we show how a practical boosting algorithm can be constructed for the $L_{2,1}^2 + \gamma L_1^2$ regularization function used in (4), by deriving the conjugate of the regularization function, $\Phi^*$, computing a sparse subgradient over finite sets, and providing a tractable oracle heuristic for boosting from infinite sets of basis vectors (Algorithm 2). For this regularization function there are always subgradients consisting of a single new basis vector at each step, and we employ a step-wise fitting approach in Algorithm 1 to boost a basis matrix for sparse coding by adding one basis vector in each boosting step. Note that either $\epsilon$ or $\lambda$ is assumed to be a known hyper-parameter. Here we assume $\lambda$ is a known constant as this leads to greater deflation between boosting steps, and a smaller, less coherent basis matrix.

### 3.3 The Regularization Conjugate $\Phi^*(Z)$

This section will prove the following lemma:

**Lemma 1** *The Fenchel conjugate of:*

$$\Phi(W) = \frac{1}{2}\|W\|_{2,1}^2 + \frac{\gamma}{2}\|W\|_1^2$$

*is the minimization over A of:*

$$\Phi^*(Z) = \inf_A \frac{1}{2}\|Z - A\|_{2,\infty}^2 + \frac{1}{2\gamma}\|A\|_\infty^2 \quad (14)$$

The infimum over the additional variable $A$ in (14) is known as the *infimal convolution* of the $L_1^2$ and $L_{2,1}^2$

terms. Since in the coding problem the dual variable $Z$ is the negative loss gradient, $Z = -\frac{1}{\lambda}\nabla L(W)$, we will see in Section 4.3 that $A$ has the effect of focusing the boosting step on the examples with the hightest loss.

Lemma 1 follows from the the fact that the Fenchel conjugate of the sum of two functions is the *infimal convolution* of their conjugates (Rifkin & Lippert, 2007). The conjugate of the $L_1^2$ squared norm term is computed from three well known facts: the dual of a squared norm $f(x) = \frac{1}{2}\|x\|^2$ is simply the square of its *dual norm*, the dual of the $L_1$-norm $\|x\|_1$ is the $L_\infty$-norm $\|z\|_\infty$, and if $f(x)$ is a squared norm multiplied by a scalar $\gamma$, then the conjugate will be multiplied by $\gamma^{-1}$, $(\gamma f(x))^* = \gamma^{-1} f^*(z)$ (Boyd & Vandenberghe, 2004)

Showing that the conjugate of the $L_{2,1}^2$ term is $L_{2,\infty}^2$, requires a lemma about the dual norm of a block norm. We establish that the dual norm of the $\|W\|_{2,1}$ block norm is $\|Z\|_{2,\infty}$, and finish the proof of Lemma 1 by proving the following lemma in (Bradley & Bagnell, 2009a), along with a version for general compositional norms:

**Lemma 2** *The dual of the $L_{p,q}$ block norm is a $L_{p^*,q^*}$ block norm where $1/p + 1/p^* = 1/q + 1/q^* = 1$, $p, p^*, q, q^* \in [1, \infty]$.*

### 3.4 The Subgradient $\partial_Z \Phi^*(Z)$

In order to apply our boosting approach (Algorithm 1) to the sparse coding problem (4), we must compute a subgradient $\phi \in \partial_Z \Phi^*(Z)$ of (14), where the dual variable $Z$ is the negative gradient of the loss: $Z = -\frac{1}{\lambda}\nabla L(W)$. Fortunately $\partial_Z \Phi^*(Z)$ is generally very sparse, and we show that it equals:

$$\frac{\partial \Phi^*(Z)}{\partial Z_j^m} = \begin{cases} 0 & \text{if } |Z_j^m| < \alpha \\ 0 & \text{if } \|\hat{Z}^m\| - \alpha\|_2^2 < \kappa \\ Z_j^m - \text{sign}(Z_j^m)\alpha & \text{otherwise} \end{cases}$$

$$\hat{\alpha} = \|\hat{A}\|_\infty$$

$$\kappa = \max_i \sum_k \left( (|Z_k^i| - \hat{\alpha})_+ \right)^2 \quad (15)$$

where $\hat{\alpha}$ is the magnitude of the largest element of the matrix $|\hat{A}|$ (derived below), which minimizes the infimal convolution in (14), and $\kappa$ is equal to the maximal squared $L_2$ norm of any row in the matrix $Z - \hat{A}$. Deriving this subgradient requires analyzing some details of the infimal convolution, but is necessary for understanding the sub-problem involved in boosting from an infinite set of basis vectors.

#### 3.4.1 Solving the Infimal Convolution

The solution $\hat{A}$ to the infimal convolution in (14) is:

$$\hat{A}_j^i = \begin{cases} Z_j^i & |Z_j^i| \leq \alpha \\ \text{sign}(Z_j^i)\alpha & |Z_j^i| > \alpha \end{cases} \quad (16)$$

where $\alpha = \|\hat{A}\|_\infty$. The infimal convolution is effectively a one-dimensional search over $\alpha \in [0, \|Z\|_\infty]$ which seeks to minimize the max of a set of piecewise quadratic functions ($\|Z^i - A^i\|_2^2, \forall i$). Replacing the $\|Z - A\|_{2,\infty}^2$ term in the infimal convolution with $\max_i \sum_j \left( (|Z_j^i| - \alpha)_+ \right)^2$ produces:

$$\Phi^*(Z) = \min_\alpha \max_i \frac{1}{2\gamma}\alpha^2 + \frac{1}{2}\sum_j \left( (|Z_j^i| - \alpha)_+ \right)^2 \quad (17)$$

Note that this optimization problem is convex with respect to $\alpha$, since it is a max over convex functions. The minimizer, $\hat{\alpha}$, can be found to any desired accuracy by an interval bisection search[8] over $\alpha \in [0, \|Z\|_\infty^2]$. The subgradient (15) is the partial derivative[9] of (17).

## 4 Oracles for Infinite Bases

The convex exemplar-based approach to clustering formulated by (Lashkari & Golland, 2008) can be applied to sparse coding to create a simple oracle from a set of exemplars. In this case, the finite set of exemplars is the set of possible basis vectors, and the subgradient (15) derived above can be found efficiently by solving the infimal convolution. However, this solution can is improved by Algorithm 2, which optimizes over an infinite set of possible basis vectors. This section defines the optimization problem that must be solved to boost from an infinite set of possible basis vectors. We start by considering the two limiting cases of $L_{2,1} + \gamma L_1$ regularization, $L_1$ regularization and $L_{2,1}$ regularization. We then present Algorithm 2 as a heuristic for finding a good solution in the general case.

### 4.1 $L_1$ Regularization

Assume the loss function is the squared reconstruction error, $L(W, B) = \frac{1}{2}\|BW - X\|_F^2$, and the regularization function is the $L_1$-norm $\Phi(W) = \frac{1}{2}\|W\|_1^2$, with conjugate $\Phi^*(Z) = \frac{1}{2}\|Z\|_\infty^2$. In this case, a subgradient of $\Phi^*(Z)$ is a matrix with one non-zero element, corresponding to a maximal element of $|Z|$. Therefore, to find the best possible basis vector (with unit $L_2$ norm) we must solve for the basis vector $b_m$ that produces the largest element of $Z$. Since $Z = -\frac{1}{\lambda}\nabla L(W) = -\frac{1}{\lambda}B^T(BW - X)$, $b_m$ is given by:

---

[8]For a fixed level of accuracy the computational complexity of this search is at most O(Nd), where $N$ is the number of examples (columns of $Z$) and $d$ is the number of rows of $Z$.

[9]One might expect that since $\alpha$ is a function of $Z_l^m$, we should be interested in the total derivative: $\frac{d\Phi^*(Z)}{dZ_j^m} = \frac{\partial \Phi^*(Z)}{\partial Z_j^m} + \frac{\partial \Phi^*(Z)}{\partial \alpha}\frac{\partial \alpha}{\partial Z_j^m}$. However, since $\frac{\partial \Phi^*(Z)}{\partial \alpha} = 0$, the total derivative is equal to the partial derivative.

$$b_m = \frac{E_m}{\|E_m\|_2}, \text{ where:} \quad (18)$$
$$E = BW - X \quad (19)$$
$$m = \arg\max_j \|E_j\|_2^2.$$

$E$ is the reconstruction loss matrix, and $m$ is the index of the example with the largest reconstruction loss. Hence $L_1^2$ regularized boosting corresponds to adding the $L_2$ projection of the loss gradient from the highest loss example to the active basis at each step.

### 4.2 $L_{2,1}$ Regularization

If instead we regularize with the $L_{2,1}^2$ block norm $\Phi(W) = \|W\|_{2,1}^2$, with conjugate $\Phi^*(Z) = \|Z\|_{2,\infty}^2$, any row $Z^m$ with maximal $L_2$ norm is a subgradient of $\Phi^*(Z)$. Hence we can optimize over all possible basis vectors $\vec{b}$, by finding the basis vector $\vec{b}_m$ best correlated with the loss gradients on all the examples (measured by $L_2$-norm):

$$\begin{aligned} b_m &= \arg\max_{\|b\|_2 \leq 1} \|b^T(BW - X)\|_2^2 \\ &= \arg\max_{\|b\|_2 \leq 1} b^T E E^T b. \quad (20) \end{aligned}$$

Again $E$ is the reconstruction error (19). Although (20) is not convex, it is well known that $b_m$ is the eigenvector associated with the maximum eigenvalue of the matrix $EE^T$ (Boyd & Vandenberghe, 2004). Running Algorithm 1 with this choice of $\Phi$ is very related to PCA, and can be interpreted as PCA with incomplete deflation.

### 4.3 $L_{2,1} + \gamma L_1$ Regularization

The $L_{2,1} + \gamma L_1$ regularization used in (4) interpolates between the behavior of $L_1$ and $L_{2,1}$ based on the value of $\gamma$. The optimal new basis vector $b_m$ will maximize the infimal convolution:

$$b_m = \arg\max_{\|b\|_2 \leq 1} \left\{ \min_\alpha \frac{\alpha^2}{\gamma} + \sum_j \left((|b^T E_j| - \alpha)_+\right)^2 \right\} \quad (21)$$

The introduction of the minimization over $\alpha$ makes (21) significantly more difficult to solve than the $L_{2,1}$ case (20). However, we can solve the infimal convolution for a finite set of candidate basis vectors, and we have already seen the solution for the limiting cases of $\gamma \to 0$ and $\gamma \to \infty$. To review: as $\gamma \to 0$, $\hat\alpha$ will converge to 0, and $b_m$ will be equal to the solution of the $L_{2,1}$ only case (20). When $\gamma \to \infty$, $\hat\alpha$ converges to $\max_j |b_m^T E_j|$, and $L_{2,1} + \gamma L_1$ regularization reduces to $L_1$ regularization. In this case[10], $b_m \propto E_j$, where $\|E_j\|_2 = \max_i \|E_i\|_2$, i.e. $b_m$ is guaranteed to be proportional to some column of the reconstruction error $E$.

---

[10] The $L_2$ projection of any column of $E$ with maximal $L_2$ norm is a valid choice for $b_m$

---

**Algorithm 2** $L_{2,1} + \gamma L_1$ Heuristic

**Input:** Scalars $\gamma$, $\eta$ and $N$, reconstruction error matrix $E \in \Re^{m \times n}$ where $E = BW - X$.
**Output:** New basis vector $b$.
**Initialization:** Compute a matrix $\tilde{B} \in \Re^{m \times N+1}$ containing candidate basis vectors as columns:
**1:** Set $\tilde{B}_1$ to the solution for $L_{2,1}$ regularization (20).
**2:** Set $\tilde{B}_2$ through $\tilde{B}_{N+1}$ equal to the $L_2$ projections of the largest (in $L_2$ norm) $N$ columns of $E$:

$$\tilde{B}_i = E_i/\|E_i\|_2, \ \forall \|E_i\|_2 > C$$

**3:** Compute the dual variable: $Z = -\frac{1}{\lambda}\tilde{B}^T E$
**4:** Find the index $m$ of a maximal row by solving the infimal convolution:

$$m = \arg\max_i \min_\alpha \frac{1}{2\gamma}\alpha^2 + \frac{1}{2}\sum_j \left((|Z_j^i| - \alpha)_+\right)^2$$

**5:** Assign $b$ to the best candidate: $b = \tilde{B}_m$.
**6:** Improve $b$ by gradient ascent using (22):
**repeat**
  **7:** $z = -\frac{1}{\lambda}b^T E$
  **8:** $b = b + \eta\frac{\partial \Phi^*(z)}{\partial b}$
**until** $\|\frac{\partial \Phi^*(z)}{\partial b}\|_2 < \epsilon$
**Return** $b$

---

Algorithm 2 provides an empirically effective method for estimating $\beta$ and $b_m$. It starts by combining the solutions for the $L_{2,1}$ and $L_1$ regularization cases discussed above into a matrix of candidate basis vectors $\tilde{B}$. Then a promising choice for $b_m$ is selected by finding the basis vector in $\tilde{B}$ which produces a maximal row of the infimal convolution over $Z = -\frac{1}{\lambda}\tilde{B}^T E$. Finally $b_m$ is improved by gradient ascent to maximize the conjugate of the regularization, $\Phi^*(Z^m)$, where $Z^m = -\frac{1}{\lambda}b_m^T E$ and the gradient is:

$$\begin{aligned} \frac{\partial \Phi^*(Z^m)}{\partial b_m} &= \frac{\partial \Phi^*(Z^m)}{\partial Z^m}\frac{\partial Z^m}{\partial b_m} \quad (22) \\ &= -\frac{1}{\lambda}\sum_j E_j \text{sign}(Z_j^m)(|Z_j^m| - \alpha)_+ \end{aligned}$$

where: $\alpha = \min_\alpha \frac{\alpha^2}{\gamma} + \sum_j \left((|b^T E_j| - \alpha)_+\right)^2$

## 5 Results on Image Denoising

We apply boosted coding (Algorithm 1) to the task of image denoising in order to evaluate its performance on a real-world task that is well-suited to sparse coding (Elad & Aharon, 2006), and lends itself particularly well to visualizing the behavior of the algorithm. The performance of alternating optimization[11] and boosted coding turn out to be quite similar on this task, with a slight advantage for the boosted approach. This result provides reassuring evidence that the non-convex but simple alternating optimization algorithm is not seriously impaired by inferior local minima on this task. Additional experimental details and results are given in the accompanying tech report to this paper (Bradley & Bagnell, 2009a).

---

[11] Used in many past works such as (Raina et al., 2007).

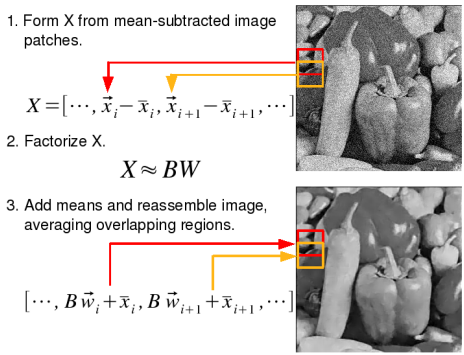

Figure 1: Image denoising proceeds by extracting overlapping patches from a noisy input image. Each patch is rearranged to form a column of the data matrix X. After X is approximated as B*W, the denoised image is reconstructed by averaging the overlapping patches.

Our approach is modeled on the K-SVD algorithm presented in (Elad & Aharon, 2006). As shown in Figure 1, overlapping patches are extracted from a noisy input image. Each patch is rearranged into a vector $\vec{x}_i$, the mean $\bar{x}_i$ of the patch is subtracted, and the result becomes a column of the data matrix $X$. $X$ is factorized into the product of $B$ and $W$ using sparse coding. Non-zero components of $W$ are then refit without regularization. Finally, the denoised image is reconstructed by adding back the mean of each patch, and averaging areas of the image where multiple patches overlap. For these experiments we used 8x8 pixel patches with an overlap of four pixels between neighboring patches. The alternating optimization approach has two hyperparameters–the regularization constant $\lambda$ and $d$, the number of columns of $B$–and boosted coding has two regularization constant hyperparameters $\lambda$ and $\gamma$. The hyper-parameters of both algorithms were independently tuned for maximal performance, in order to isolate the effect of the different optimization strategies.

Independent and identically distributed Gaussian noise ($\sigma = 0.1$) was added to five common benchmark images to match the assumed noise model of the $L_2$ loss function (Table 1, left-most column). Subtracting the mean of each patch removes low-frequency components of the image, leaving the coding problem to focus on identifying which high-frequency components of the image are contained in each patch. If $W$ is set to zero, the result is to average each 8x8 patch of the image, which improves the Signal-to-Noise Ratio (SNR) of the low-frequency components of the image at the expense of the high-frequency details in the image (Table 1, second column).

| Image | Input | P.A. | Alt. Opt. | B.C. |
|---|---|---|---|---|
| Barbara | 20.00 | 21.99 | 28.19→28.35 | 28.35 |
| Boat | 20.02 | 22.96 | 28.37→28.43 | 28.44 |
| Lena | 20.02 | 25.02 | 30.24→30.31 | 30.32 |
| House | 20.01 | 23.74 | 31.74→31.85 | 31.88 |
| Peppers | 19.97 | 21.01 | 29.04→29.09 | 29.20 |

Table 1: Results on five benchmark images show that both alternating optimization (Alt. Opt.) and boosted coding (B.C.) produce similar results for image denoising, and improve significantly on patch averaging (P.A.). The range reported for alternating optimization is the best and worst performance from 20 randomly initialized trials. Note that boosted coding is deterministic.

Coding $X$ with alternating optimization or boosted coding (Table 1, right side) restores high-frequency detail by finding basis vectors that describe shared patterns across all patches. In our experiments both algorithms produce roughly equivalent results in terms of SNR, with a slight advantage for the convex approach.

### 5.1 Boosted Coding

The effect of relaxing the non-convex rank constraint on $W$ by substituting $L_{2,1}$ regularization changes the basis vectors selected. Boosted coding selects the most important basis vectors first, and those are used by many image patches due to the "group discount" provided by the $L_2$ norm in $L_{2,1}$ regularization (Figure 2). This causes the signal to noise ratio to rise quickly at the beginning of the process and then level off once most of the underlying signal can be represented by the basis vectors.

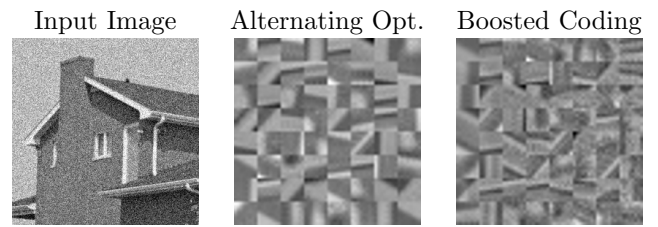

Figure 2: **Left:** Noisy input image. **Center:** 8x8 basis vectors learned by alternating optimization. **Right:** Basis vectors learned by boosted coding, displayed in the order they were selected (top to bottom and left to right).

The first basis vectors chosen are smoother in appearance than basis vectors chosen at later steps. This is because each step of boosting with $L_{2,1} + \gamma L_1$ regularization will find a basis vector that is maximally correlated with the reconstruction error on a subset of the image patches. In later rounds of boosting much of the structure of the image patches is already explained, and the reconstruction error on each patch consists largely of noise. Additionally, the basis selected by boosted coding is less coherent (i.e. the basis vectors are less correlated with each other) than the basis selected by alternating optimization (details in (Bradley & Bagnell, 2009a)).

### 5.2 Alternating Optimization

A common objection to the traditional, alternating optimization approach to sparse coding (2) is that the non-convex rank constraint on $B$ could result in the algorithm returning inferior local minima. Anecdotal evidence suggests this problem should be most acute for relatively small basis sizes, where there are only a small number of randomly-initialized basis vectors. In this case there are fewer degrees of freedom available to let alternating optimization escape a local minima. In our experience, inferior local minima, while they do occur, have a relatively small effect on the performance of the alternating optimization algorithm. Table 2 quantifies this assertion by showing the results of repeatedly running the alternating optimization image denoising algorithm from different random initializations of the basis vectors. The optimization alternated between $B$ and $W$ 20 times. This represents a stress-test for the alternating optimization algorithm as the basis contains only eight basis vectors. Even in this challenging case alternating optimization performs reasonably consistently, although Figure 3 provides a detailed look at a case where one local minima was superior to the others.

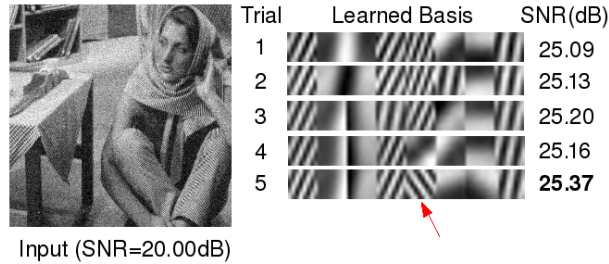

Figure 3: **Left:** input image. **Right:** five bases learned by alternating optimization starting from random initializations. The basis that performed best found an important pattern (indicated) that was not representable in the other bases, which were stuck in worse local minima.

| Image | SNR Improvement (dB) | | |
|---|---|---|---|
| | Min | Max | Mean |
| Lena | 4.60 | 4.78 | $4.73 \pm 0.06$dB |
| Barbara | 3.10 | 3.38 | $3.20 \pm 0.11$dB |
| Boat | 4.57 | 4.65 | $4.62 \pm 0.03$dB |
| Peppers | 6.55 | 6.84 | $6.71 \pm 0.1$dB |
| House | 6.20 | 6.25 | $6.22 \pm 0.02$dB |

Table 2: Signal-to-Noise ratio variance observed when using alternating optimization $L_1$-regularized sparse coding to denoise grayscale images from 20 different random initializations of B.